\documentclass[10pt,twocolumn,letterpaper]{article}

\usepackage{abstract}
\usepackage{iccv}
\usepackage{times}
\usepackage{epsfig}
\usepackage{graphicx}
\usepackage{amsmath}
\usepackage{amssymb}
\usepackage{multirow}
\usepackage{dsfont}
\usepackage[dvipsnames]{xcolor}
\usepackage{subcaption}
\usepackage{svg}
\usepackage{nicefrac}
\usepackage{float}

\usepackage[pagebackref=true,breaklinks=true,letterpaper=true,colorlinks,bookmarks=false]{hyperref}

\iccvfinalcopy 

\def\iccvPaperID{7581} 

\ificcvfinal\fi

\begin{document}

\title{SemiGPC: Distribution-Aware Label Refinement for Imbalanced Semi-Supervised Learning Using Gaussian Processes}


\author{Abdelhak Lemkhenter\thanks{Currently at University of Bern. Work conducted during an internship at Amazon.}, $\;$ Manchen Wang \thanks{Corresponding author.}, $\;$ Luca Zancato, $\;$ \\
Gurumurthy Swaminathan, $\;$ Paolo Favaro, $\;$ Davide Modolo\\ 
AWS AI Labs \\
{\tt\small abdelhak.lemkhenter@inf.unibe.ch, $\;$ \{manchenw, zancato, gurumurs, pffavaro, dmodolo\}@amazon.com}}


\twocolumn[{%
\renewcommand\twocolumn[9][]{#1}%
\maketitle
    \setkeys{Gin}{width=0.196\linewidth}
    \renewcommand\thesubfigure{}
    \setlength\tabcolsep{0.5pt}

\centering
\setlength{\tabcolsep}{20pt}
\begin{tabular}{ccc}
    \includegraphics[width=0.25\textwidth]{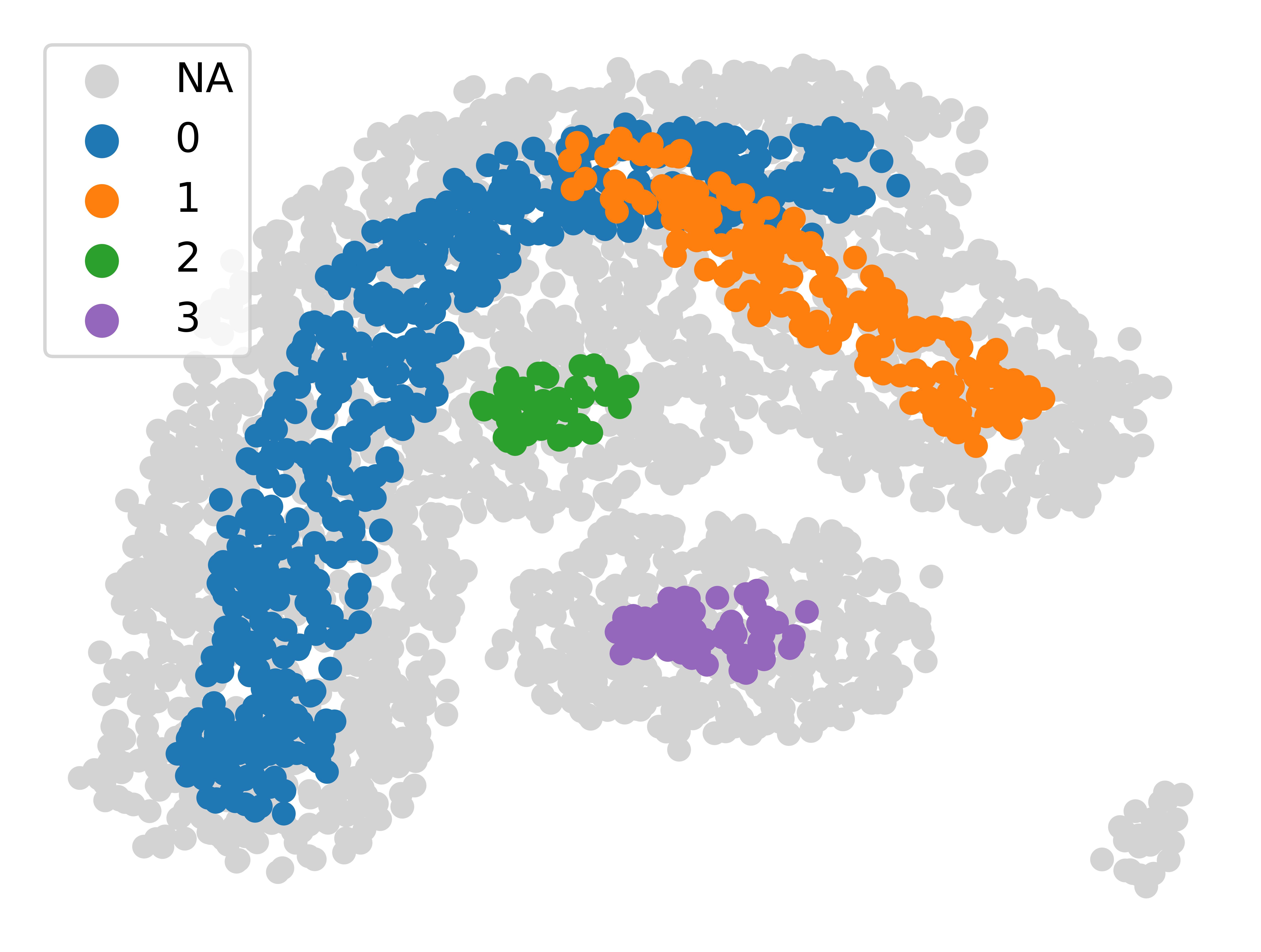} &  
    \includegraphics[width=0.25\textwidth]{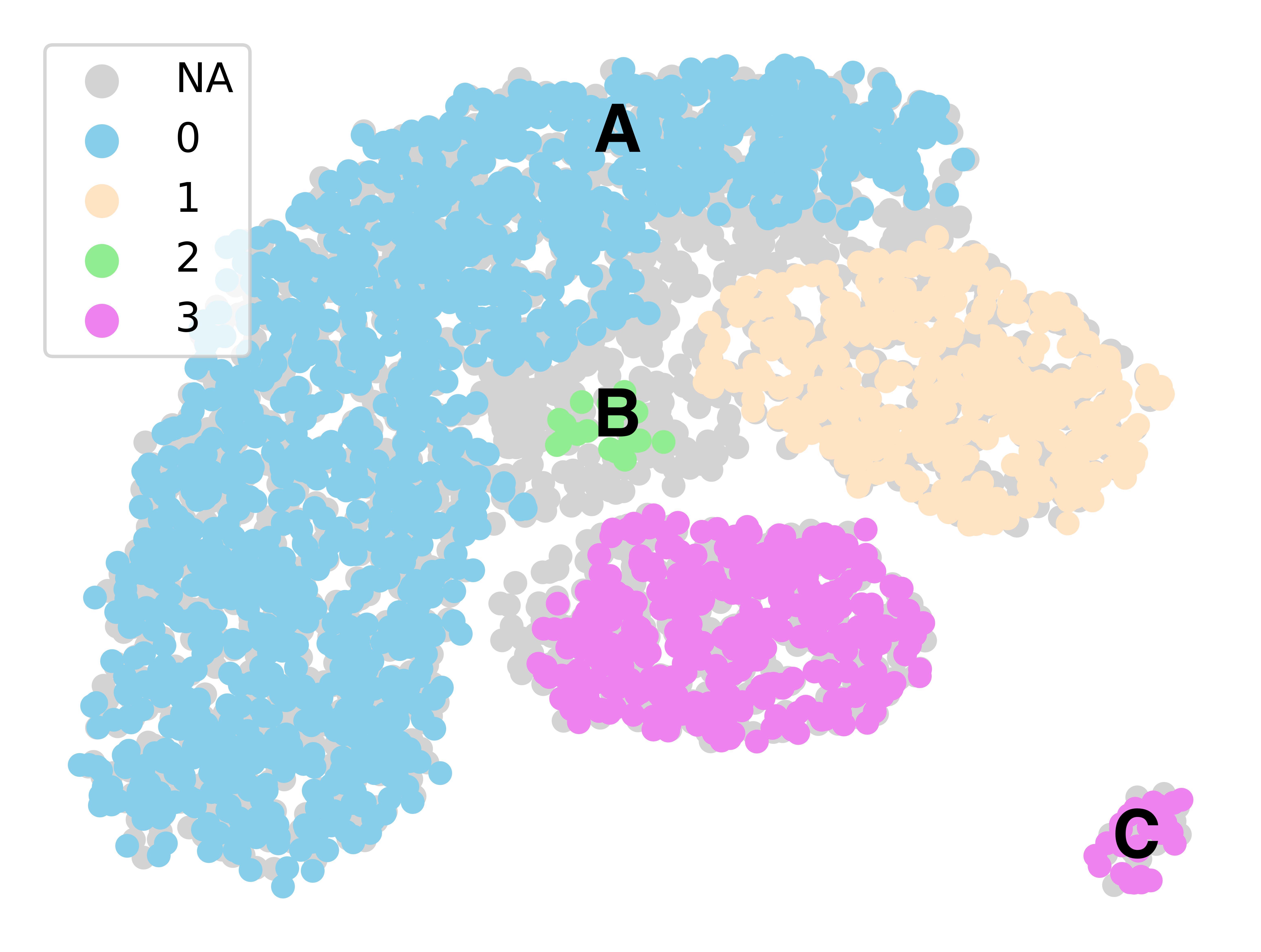} & 
    \includegraphics[width=0.25\textwidth]{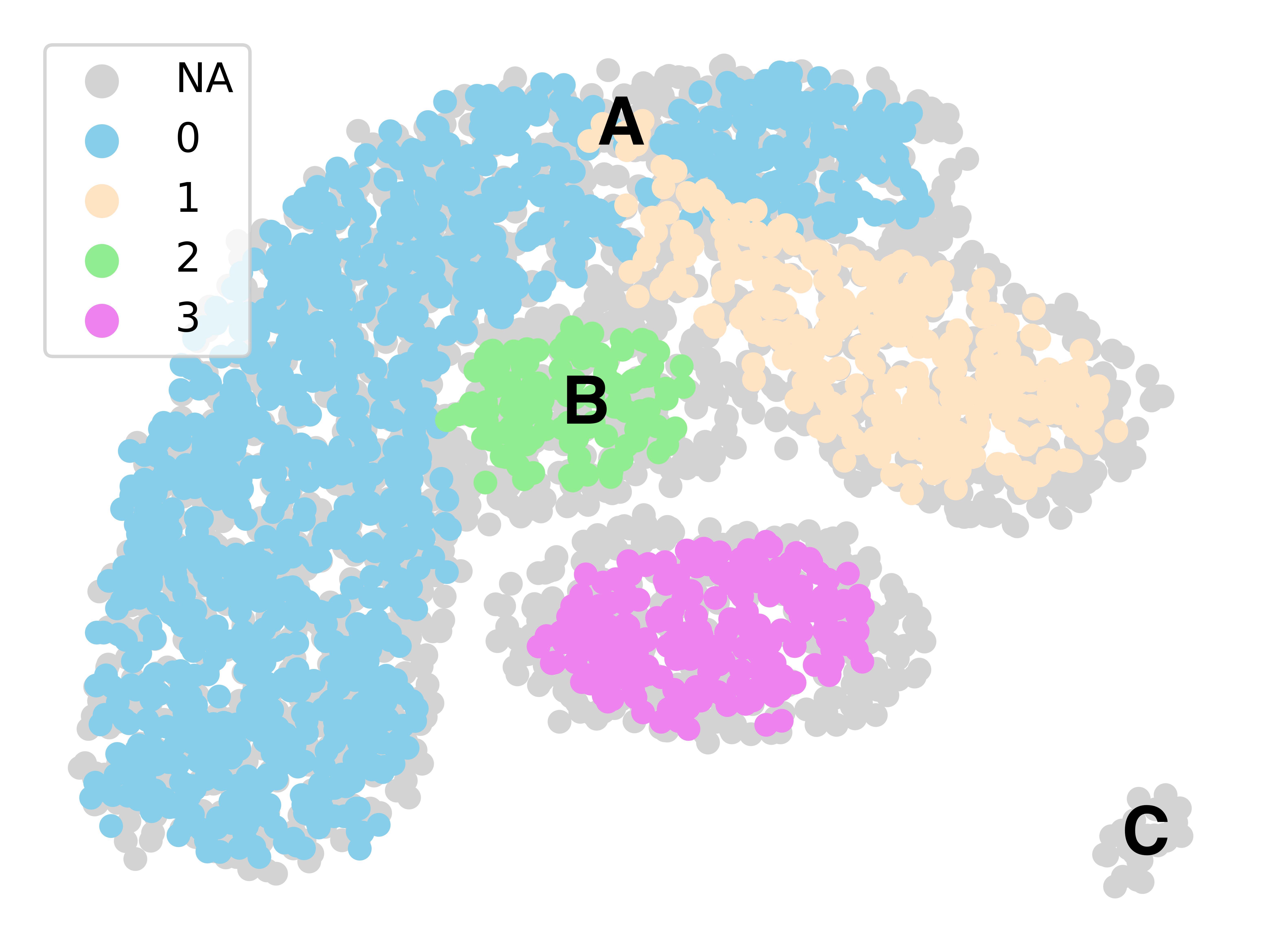} \\
    (a) Initial labels &
    (b) Similarity-based predictions &
    (c) SemiGPC predictions 
\end{tabular}

\captionof{figure}{\label{fig:label-ref} {\bf SemiGPC pseudo-labeling with class imbalance}. (a) Four-class dataset containing unlabeled (gray) and labeled samples (in color). (b) and (c) show the labels propagated according to the similarity-based aggregate~\cite{li2021comatch} and SemiGPC methods. (c) In A, the initial labels are mixed so SemiGPC is more conservative there (many samples are not pseudo-labeled at the current threshold level). In B, SemiGPC is able to propagate the labels of the minority green class despite being surrounded by the majority blue one. In C, SemiGPC assigns low confidence to the set of outliers (labels are not propagated). In contrast, the similarity-based approach expands the majority classes at the expense of the minority ones, \cf B and C. \\
}
}]

\ificcvfinal\thispagestyle{empty}\fi


\begin{abstract}
   In this paper we introduce SemiGPC, a distribution-aware label refinement strategy based on Gaussian Processes where the predictions of the model are derived from the labels posterior distribution. Differently from other buffer-based semi-supervised methods such as CoMatch~\cite{li2021comatch} and SimMatch~\cite{zheng2022simmatch}, our SemiGPC includes a normalization term that addresses imbalances in the global data distribution while maintaining local sensitivity. This explicit control allows SemiGPC to be more robust to confirmation bias especially under class imbalance.
   We show that SemiGPC improves performance when paired with different Semi-Supervised methods such as FixMatch~\cite{sohn2020fixmatch}, ReMixMatch~\cite{berthelot2020remixmatch}, SimMatch~\cite{zheng2022simmatch} and FreeMatch~\cite{wang2023freematch} and different pre-training strategies including MSN~\cite{assran2022msn} and Dino~\cite{caron2021dino}.
   We also show that SemiGPC achieves state of the art results under different degrees of class imbalance on standard CIFAR10-LT/CIFAR100-LT especially in the low data-regime. 
   Using SemiGPC also results in about 2\% avg.~accuracy increase compared to a new competitive baseline on the more challenging benchmarks SemiAves, SemiCUB, SemiFungi~\cite{su2021realistic} and Semi-iNat~\cite{su2021tax}. \\
   
\end{abstract}

\saythanks


\section{Introduction}

Semi-Supervised Learning offers a more cost effective alternative to fully supervised learning when scaling up the data collection process.  
Current state the of the art semi-supervised methods rely on self-learning by generating pseudo-labels for the unlabeled samples. However, pseudo-labels can also hurt the final performance when they introduce persistent incorrect predictions, a problem known as confirmation bias.  In particular, self-learning can bias the label distribution if the data is imbalanced. To address this, recent works such as CoMatch~\cite{li2021comatch} and SimMatch~\cite{zheng2022simmatch} rely on a buffer of samples to refine the predicted pseudo-labels.
However, no counter measure is adopted to globally balance the data in the memory bank. As such, the resulting \textit{refined pseudo-labels} are plagued by the class imbalances present in the unlabeled data.



To overcome these limitations we introduce SemiGPC, a novel semi-supervised learning method that generates pseudo-labels using a distribution-aware label-refinement strategy. This distribution awareness stems from the use of Gaussian Processes, which accounts for local data concentration disparities and counteracts them. This results in more robust  pseudo-labels especially for minority classes and outliers as shown in Figure~\ref{fig:label-ref}. In particular, SemiGPC correctly assigns nearby points to the minority classes despite the larger count of the majority class at a bigger scale, \ie it has a better local sensitivity, while remaining faithful to the global data distribution. 
SemiGPC is flexible and can be used on-top of previous label-refinement schemes on other semi-supervised methods. Furthermore, we show that the similarity-based pseudo-labels heuristics used in SimMatch and CoMatch can be cast as a special case of SemiGPC.
To improve computational efficiency of our method and allow for fast batched updates essential for Semi-Supervised learning methods, we pair SemiGPC with a batched online update rule that significantly reduces its forward pass cost ($\times 7.5$ speed-up).
We show the benefit of using SemiGPC on top of semi-supervised algorithms such as FixMatch~\cite{sohn2020fixmatch}, ReMixMatch~\cite{berthelot2020remixmatch}, SimMatch~\cite{zheng2022simmatch} and FreeMatch~\cite{wang2023freematch} ($\sim$ 0.8\% avg. improvement) and different self-supervised pre-training strategies such as MSN~\cite{assran2022msn} and Dino~\cite{caron2021dino} resulting in a $\sim$ 1.3\% avg. improvement. This highlights the general purpose nature of SemiGPC as a relevant extension for semi-supervised methods based on label refinement strategies. We experimentally show that SemiGPC is able to achieve state of the art results on CIFAR10-LT ($\geq +7.65\%$) and CIFAR100-LT ($\geq +1.84\%$) as well as the more challenging semi-supervised benchmarks SemiAves, SemiCUB, SemiFungi and Semi-iNat ($\sim +1.92\%$ compared to our baseline and $\sim + 20.52\%$ compared to numbers reported in the literature~\cite{su2021realistic, su2021tax}). We also show that SemiGPC is able to narrow the gap between the high and low data regimes with 10-100x fewer labeled samples as we report a $45\%$ and $32\%$ relative improvement over the baseline across regimes for CIFAR10-LT and CIFAR100-LT respectively.

\section{Related Works}

\noindent \textbf{Consistency-based Semi-Supervised Learning.}
Semi-supervised learning methods such as FixMatch~\cite{sohn2020fixmatch}, ReMixMatch~\cite{berthelot2020remixmatch}, SimMatch~\cite{zheng2022simmatch} and FreeMatch~\cite{wang2023freematch} share the common design choice of enforcing the consistency of the model predictions across augmentations of different strength levels on the unlabeled samples. FixMatch~\cite{sohn2020fixmatch} enforces this consistency as a cross-entropy loss applied using the one-hot encoding of the model predictions on weakly augmented unlabeled samples as pseudo-labels for  their strongly augmented counterpart. Weak augmentations consist of random image flipping and translation while strong augmentation combine AutoAugment~\cite{cubuk2019autoaugment} and Cutout~\cite{devries2017cutout}.
The consistency loss is only applied on high confidence predictions where the predicted probability value is higher than a fixed threshold. ReMixMatch~\cite{berthelot2020remixmatch} instead opts for using temperature sharpened model predictions as pseudo-labels for its consistency regularization in addition to a rotation prediction regularization loss. 
FreeMatch~\cite{wang2023freematch} introduces a per-class confidence threshold update rule based on the models predictions combined with an entropy-based diversity loss making it more suitable for imbalanced settings. 
Methods such as CoMatch~\cite{li2021comatch} and SimMatch~\cite{zheng2022simmatch} choose to enforce consistency across additional data representations. In particular, CoMatch~\cite{li2021comatch} encourages the consistency between the pseudo-labels and embeddings similarity graphs while SimMatch~\cite{zheng2022simmatch} encourages the consistency between semantic-level and instance-level pseudo-labels. Both methods choose to smooth the predicted pseudo-labels based on a similarity based aggregate of a memory buffer of samples in order to mitigate confirmation bias. However, these pseudo-label refinement strategies fail to eliminate data biases \wrt the class balance.

\noindent \textbf{Probabilistic Models for Semi-Supervised Learning.} Gaussian Processes (GP) are a class of non-parametric function approximation methods fully characterized by their mean and kernel functions. Given a set of observations and their corresponding measurements, GPs define a posterior distribution over measurements for new observations. Their ability to explicitly model uncertainty makes them a natural fit for Semi-supervised learning. Early works such as~\cite{lawrence2004semi} introduce GPs in the context of Semi-supervised learning by assuming that the data density in regions between the class-conditional densities should be low while ~\cite{sindhwani2007semi} leverages GPs to model the relationship between labeled and unlabeled samples by incorporating the geometry of the latter in the construction of the global kernel function. However such early works are limited to toy datasets due to the computational cost of GP. The more recent UaGGP work~\cite{liu2020uncertainty} proposed to address uncertainty caused by erroneous neighborhood relationships in the context of graph-based semi-supervised learning by leveraging the ability of GPs to generalize well from few samples.
NP-Match~\cite{wang2022np} proposed Neural Processes instead of GPs as a probabilistic model for uncertainty estimation which, in turn, allows for better computational efficiency compared to MCDropout~\cite{gal2016dropout}. 

\noindent \textbf{Gaussian Processes and Deep Learning.} Beyond Semi-Supervised Learning, Gaussian Processes have been used alongside neural networks in multiple other fields. DGPNet~\cite{johnander2022dense} relies on GPs in the context of dense few-shot segmentation to capture complex appearance distributions while \cite{pmlr-v164-lee22c} leverages GPs for fast and accurate uncertainty estimates in robotics systems. Furthermore, different works draw parallels between Gaussian Processes and Neural Networks by interpreting the action functions of the latter as as interdomain inducing features~\cite{dutordoir2021deep} or by proving a correspondence between the two classes of models \cite{yang2019wide}.
\section{SemiGPC}\label{sec:methods}


\begin{figure}[t!]
    \centering
    \includegraphics[width=0.45\textwidth]{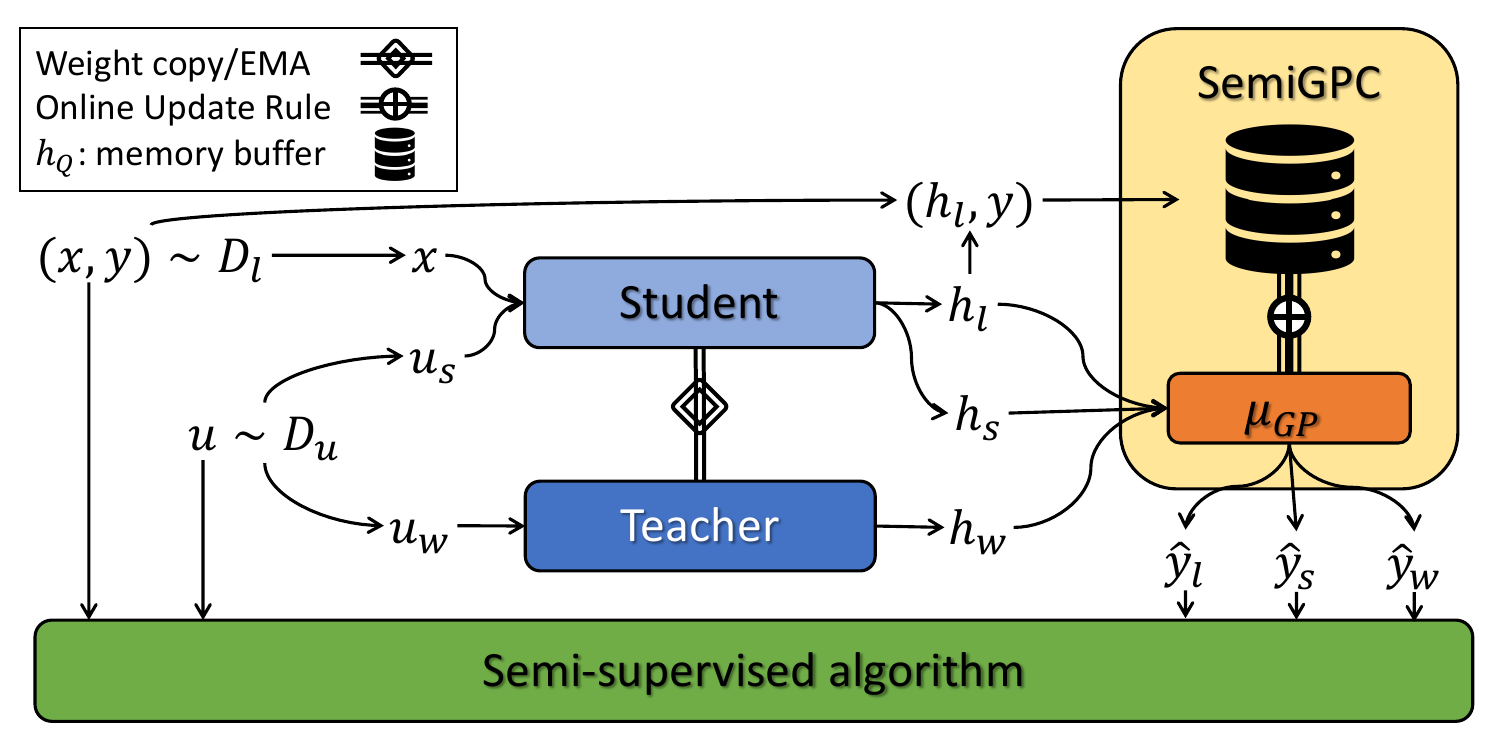}
    \caption{{\bf SemiGPC Outline.} $x,u_s,u_w,h$ and $y$ are the labeled, strongly/weakly augmented unlabeled samples, their feature vector and the ground truth labels respectively. The SemiGPC buffer is used to derive the model predictions~  
    $\hat{y}$.}
    \label{fig:semigpc}
\end{figure}

In this section, we present the basic framework of consistency-based semi-supervised learning, how it can be extended with a memory buffer and analyze where confirmation bias comes into play. We then introduce our Gaussian Processes-based classifier, SemiGPC, and highlight its advantages over other classifiers using a toy example. 
The general outline of SemiGPC is shown in Figure~\ref{fig:semigpc}. 
In the following, we shall denote the labeled dataset with $\mathcal{D}_l = \{(x_l^i, y_l^i)\}_{i=1}^{n_{l}}$ and the unlabeled one with $\mathcal{D}_u = \{(x_u^i)\}_{i=1}^{n_{u}}$, where $x \in X$ are RGB images and $y \in \mathbb{R}^C$ are labels belonging to a fixed set of concepts $C$. We indicate a feature extractor with $h: X \to Z$ where $Z=\mathbb{R}^d$ and $d$ is the dimension of the feature space, and call the classification head $g: Z \to \mathbb{R}^C$. The model predictions are defined as $\hat{y}(x) = g \circ h(x)$.

\subsection{Consistency-based Semi-Supervised Learning}
Given labeled and unlabeled datasets $\mathcal{D}_l$ and $\mathcal{D}_u$, consistency based Semi-Supervised methods~\cite{sohn2020fixmatch, berthelot2020remixmatch, zheng2022simmatch, li2021comatch, wang2023freematch} 
rely on the labeled set and high-confidence pseudo-labels computed on the unlabeled set. 
Pseudo-labels are computed using strongly $\text{aug}_s(x)$ and weakly $\text{aug}_w(x)$ augmented views of a given image $x$, as follows: $\hat{y}_s(x) = g(h(\text{aug}_s(x)))$ and $\hat{y}_w(x) = g(h(\text{aug}_w(x)))$. We refer to \cite{sohn2020fixmatch, li2021comatch, zheng2022simmatch} for typical strong and weak data augmentations. In this work, we adopt those used in FixMatch~\cite{sohn2020fixmatch}.
More precisely, the labeled and unlabeled losses are defined as 
\begin{align}
    \mathcal{L}_l &= H(\hat{y}_w(x), y);\quad  (x, y) \in \mathcal{D}_l \nonumber \\
    \mathcal{L}_u &= \mathds{1}_{[\text{conf}(x) > \tau]} H(\hat{y}_s(x), f(\hat{y}_w(x)));~~  x \in \mathcal{D}_u \nonumber \\
    \text{with } &\text{conf}(x)=\max[\text{softmax}(\hat{y}_w(x))] \label{eq:conf}
\end{align}
where $H$ is the cross-entropy loss, $\tau$ is the confidence threshold specifying which unlabeled samples to use and $f: \mathbb{R}^C \to \mathbb{R}^C$ is a label refinement function (\eg see~\cite{berthelot2020remixmatch}). The model confidence is defined as the maximum of the softmax vector. 
Different choices of $f$ include the identity function (no refinement), one hot encoding (hard pseudo-labels used in FixMatch~\cite{sohn2020fixmatch}), temperature sharpening~\cite{berthelot2020remixmatch}, \etc. 

For a linear classification head, such pseudo-labels are sensitive to outliers in the sense that a new unlabeled sample located far from the labeled data can have high confidence, \cf Figure~\ref{fig:lin-conf-imb}. To overcome such limitation, works such as SimMatch~\cite{zheng2022simmatch} and CoMatch~\cite{li2021comatch} propose to ground their pseudo-labels using a memory buffer during training. The buffer, $(h_Q, y_Q)$, is a set of $N_Q$ feature vectors of weakly augmented labeled samples using $\text{aug}_w$, \ie, $h_Q := \{ h(\text{aug}_w(x_l)); x_l \sim \mathcal{D}_l\} \in \mathds{R}^{n \times d}$, and their associated labels.
Then, the smoothed pseudo-labels (output of $f$) on any given input $x$ are defined as 
\begin{align}
    f(\hat{y}(x))) =& \quad (1-\alpha) \hat{y}(x) + \alpha \hat{y}^{sim}  \\
    \text{with }   \hat{y}^{sim} =& \quad k(h(x), h_Q) y_Q \label{eq:sim_pl}
\end{align}
where $\alpha$, $k$ and $h(x)$ are a smoothing factor, a kernel similarity function and the feature representation of the input $x$.
The pseudo-labels $\hat{y}^{sim}$ introduced in eq.~\eqref{eq:sim_pl} closely reflects the data distribution.
However, biases present in the data, if not addressed, could be amplified due to the un-weighted kernel average (\eg, by favoring the majority classes \cf Figure~\ref{fig:sim-conf-imb} over minority ones).
In this work, we address short-comings of previous approaches by introducing a normalization term, computed leveraging Gaussian Processes, that automatically counteracts class imbalances in the data.

\subsection{Gaussian Processes-based Label Refinement}\label{sec:gp-label-refinement}

\begin{figure*}[h!]
    \centering
    \begin{subfigure}[t]{0.33\textwidth}
        \centering
        \includegraphics[width=\textwidth]{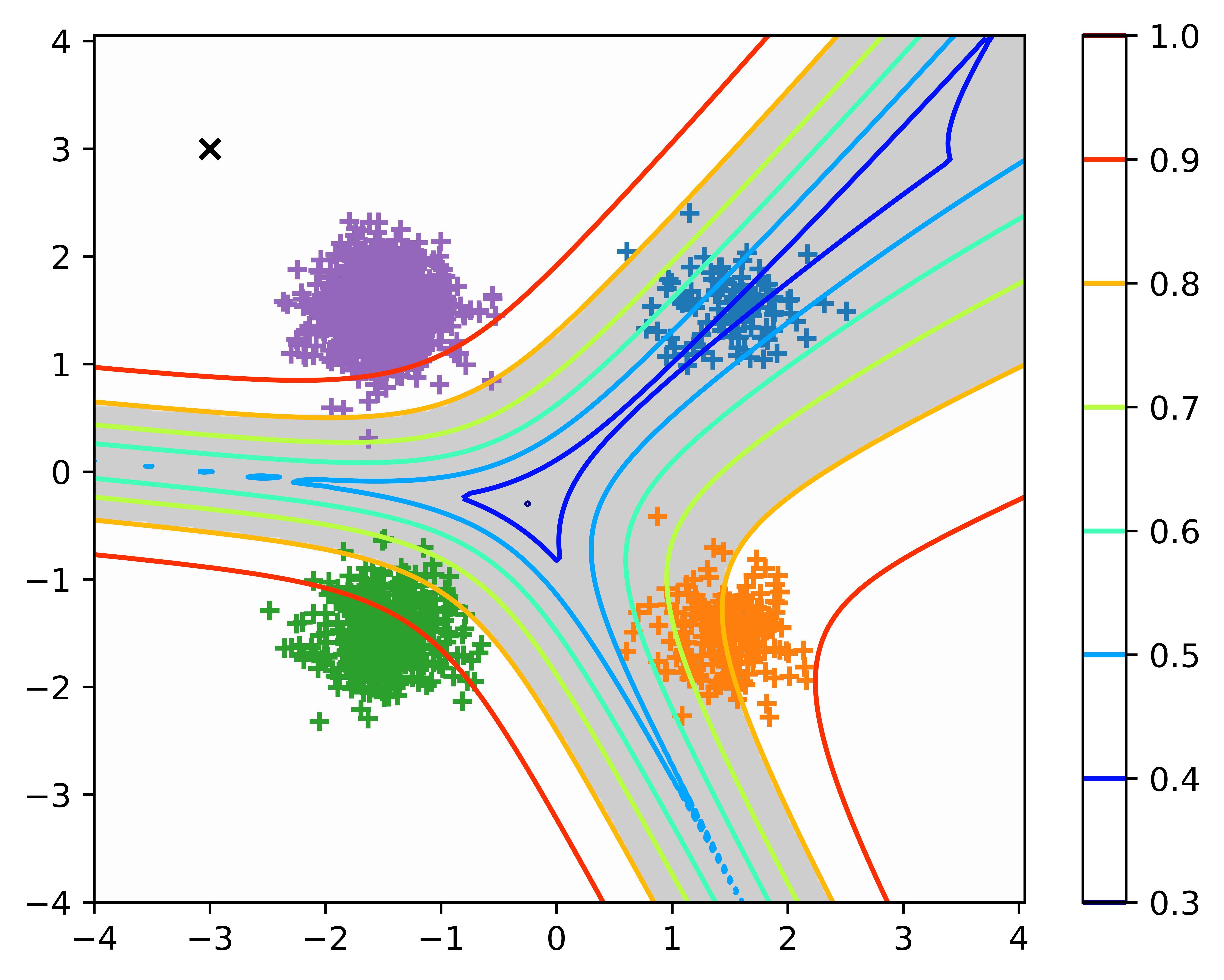}
        \caption{Linear classifier confidence}
        \label{fig:lin-conf-imb}
    \end{subfigure}%
    ~ 
    \begin{subfigure}[t]{0.33\textwidth}
        \centering
        \includegraphics[width=\textwidth]{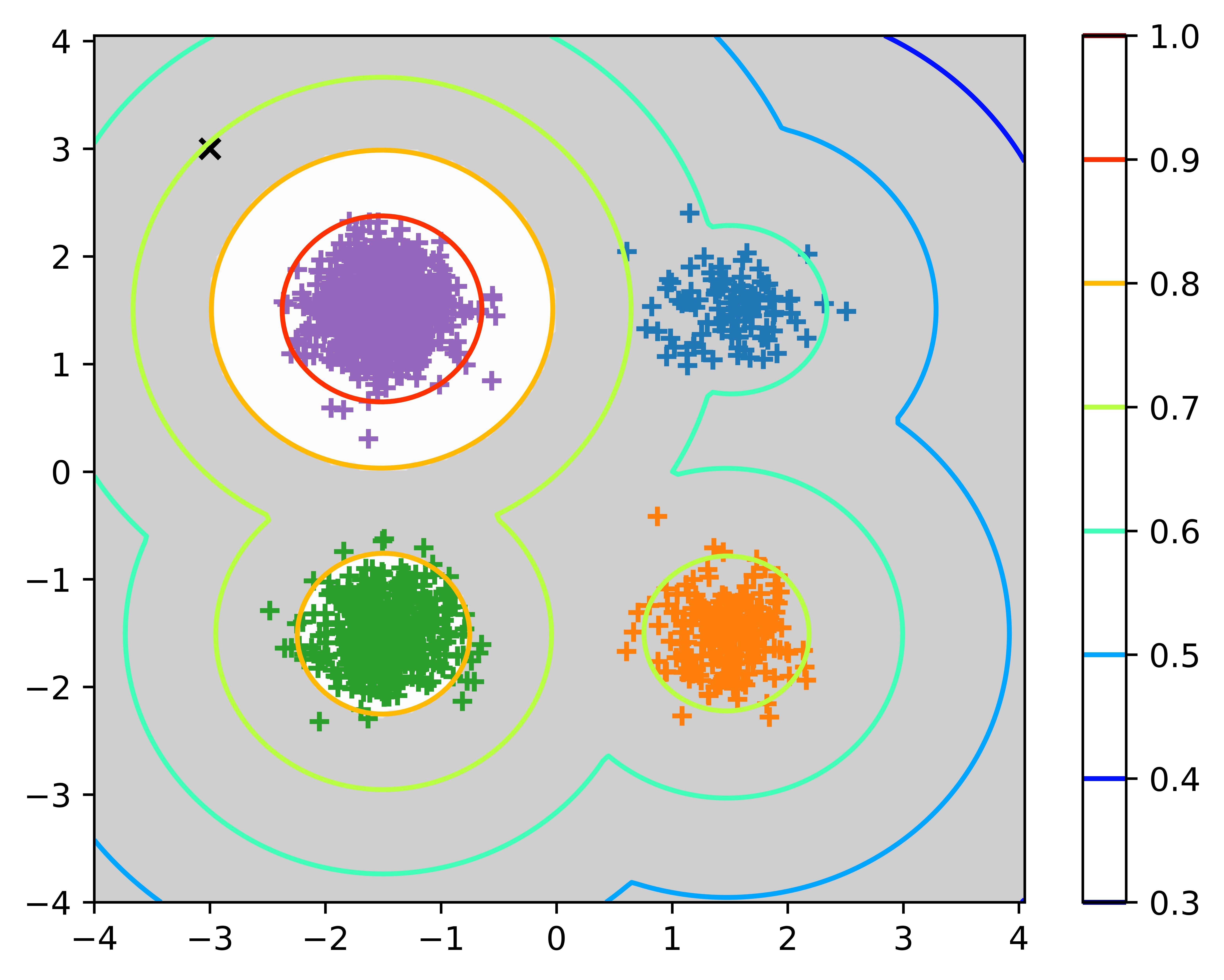}
        \caption{Similarity-based classifier confidence}
        \label{fig:sim-conf-imb}
    \end{subfigure}%
    ~ 
    \begin{subfigure}[t]{0.33\textwidth}
        \centering
        \includegraphics[width=\textwidth]{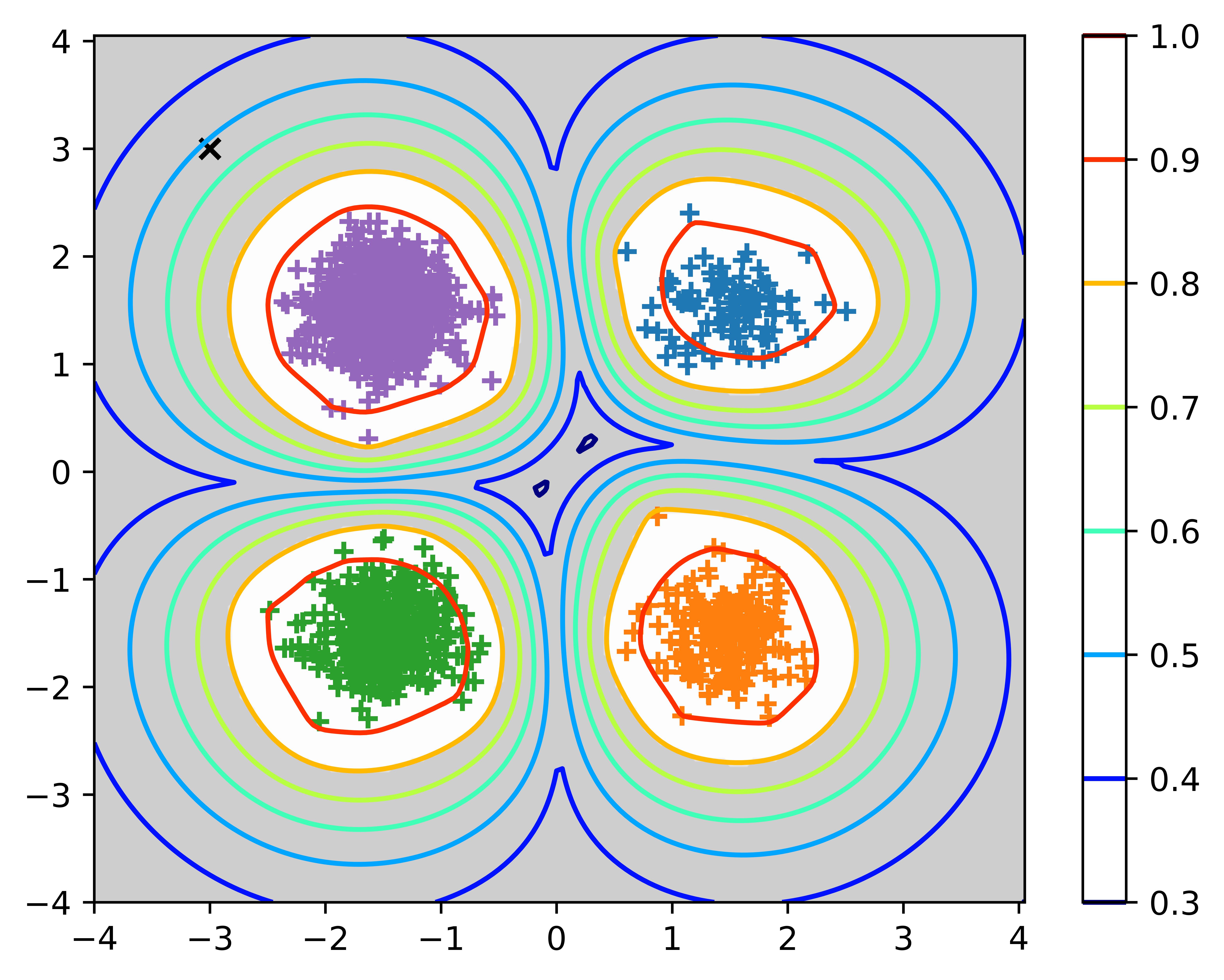}
        \caption{GP classifier confidence}
        \label{fig:gp-conf-imb}
    \end{subfigure}
    \caption{{\bf Comparison of confidence maps}: (a) a linear model, (b) a similarity-based classifier~\cite{li2021comatch} and (c) a GP classifier are represented using the contour lines. The number of samples per class grows clockwise by a factor of 2 starting from the top right cluster. We grey out regions that are below 80\% confidence. The outlier at (-3,3) is indicated with an $\times$. (c)~Only the GP classifier is able to define confidence levels that are not biased toward the majority classes and ignore the outlier.}
    \label{fig:uncertainty}
\end{figure*}

We now introduce SemiGPC, our label refinement strategy based on Gaussian Processes (GPs). Our key motivation is that the normalized kernel similarity used in the GP posterior mean helps address class imbalance by equalizing the local contribution of each sample in the buffer \wrt each class population.
Similarly to previous methods \cite{sohn2020fixmatch, li2021comatch, zheng2022simmatch}, SemiGPC aggregates global information for the refinement of each pseudo-label's input-location. However, SemiGPC retains local sensitivity by favoring the minority classes when appropriate despite the global aggregate favoring the majority ones as first shown in Figure~\ref{fig:label-ref}.  

Given a memory buffer containing features and labels $(h_Q, y_Q)$, we define SemiGPC refined pseudo-labels as
\begin{align}
    \hat{y}^{GP}(x) = \lambda~\mu^{GP}(h(x)) &= \lambda~k(h(x), h_Q) K^{-1} y_Q \label{eq:gp-pl}\\
    \text{with } K &= k(h_Q, h_Q) + \sigma ^ 2 I \nonumber
\end{align}
where $\mu^{GP}$ is the posterior mean of the GP, $\lambda$ is the logit scaling factor, $\sigma$ a regularization parameter of the GP which represents how much we trust labels in the memory bank and $k$ is the GP kernel function (\eg, the RBF kernel). 
By comparing equations~\eqref{eq:gp-pl} and~\eqref{eq:sim_pl} we see that the GP approach aggregates all labels in the memory bank and re-weights them according to the inverse covariance matrix $K^{-1}$. Such normalization is particularly useful to counteract class imbalance as we show in Figure~\ref{fig:uncertainty}.

In the following, we use the RBF kernel defined as $k(x,y) = \eta \exp(- \nicefrac{\|x -y\|^2}{2l^2})$ where $\eta$ and $l$ are the kernel scale factor and length scale respectively. 
Note that eq.~\eqref{eq:gp-pl} characterizes the posterior mean of a GP whose likelihood function is Gaussian, in general, other non-Gaussian choices are available and typically applied to build GP-based classifiers \cite{rasmussen2006gaussian}. However, when non-Gaussian likelihood are used, no closed-form solution exists and approximation schemes which entail higher computational costs are required \cite{rasmussen2006gaussian}. Thus, we choose to refine pseudo-labels by directly regressing the logits $\hat{y}$ using a Gaussian likelihood.

\noindent \textbf{Connection with other label refinement methods}.
Eq.~\eqref{eq:gp-pl} can also be rewritten as $\mu^{GP}(x)=k(h(x), h_Q) y_{K}$, a similarity-based aggregation of $y_K$, the propagated version of $y_Q$ through the graph defined by $K$.
When $\nicefrac{\eta}{\sigma^2} \to 0$ $K \to \sigma ^2 I$. In this setting, eq.~\eqref{eq:gp-pl} becomes equivalent to eq.~\eqref{eq:sim_pl}. Thus, we obtain the similarity-based aggregation strategy of works such as SimMatch~\cite{zheng2022simmatch} and CoMatch~\cite{li2021comatch}. Furthermore, clipping the kernel below a given threshold results in a matrix $K$ equivalent to 
an epsilon graph.

\noindent \textbf{SemiGPC robustness to class imbalance}.
We now illustrate with a toy example how SemiGPC is more robust to class imbalance than previous methods. In particular, we compare SemiGPC with a linear classifier and the similarity-based classifier used in CoMatch~\cite{li2021comatch} in Figure~\ref{fig:uncertainty}. We build a dataset with 4 normally distributed classes centered at (1, 1), (1, -1), (-1, -1) and (-1, 1) resp., and plot the model \textit{confidence} as defined in eq.~\eqref{eq:conf}. To simulate class imbalance, the number of samples per class grows by a factor of 2 starting from the top right cluster and going clock-wise. 

First, note how samples far from the data distribution, \eg (-3, 3), are assigned very high confidence by the linear model although such points are locally isolated from the others and therefore should not be considered well supported by evidence. 
Second, note that the minority class (in blue) is a low confidence region for both the linear and similarity-based classifier, despite locally containing many samples supporting that class. 
On the other hand, the GP-based classifier defines an appropriate high confidence region for each supported class and its sensitivity to the confidence threshold is much smaller than the similarity-based classifier used in CoMatch~\cite{li2021comatch} as highlighted by the contour plots.

Summarizing, thanks to the use of a GP-based label refinement strategy, SemiGPC is confident if: {(1)} the considered sample is close to a subset of $h_Q$ regardless of whether it belongs to the majority or minority classes since $K^{-1}$ reweighs the kernel similarity to counteract disparities in class populations while all samples far from the data are considered outliers, and {(2)}  the sample is located in a high purity region \wrt $y_Q$ since the average of conflicting results in a model confidence that is spread between classes.

\subsection{Efficient GP update}\label{sec:fast-gp}
As we mentioned in section~\ref{sec:gp-label-refinement}, applying GPs in a classification setting requires approximation schemes that are in general computational expensive. To reduce the forward time of SemiGPC we choose to model the refined pseudo-labels using a Gaussian likelihood. In this way, computing the posterior mean for each input image only requires solving a quadratic optimization problem available in closed-form.
However, computing $\mu^{GP}$ is still computationally expensive since we need to update the set $h_Q$ after each model update and invert the covariance matrix $K$ which scales with the cube of the memory bank size ($N_Q$) at each mini-batch forward pass.
To speed up computations we start from the key observation that at each optimization iteration, most of the samples in the queue do not change. Therefore, at the $t$-th iteration after observing the new batch of data of size $B$, we update the previously computed covariance at step $t-1$ with an incremental update rule. In the following, we implement SemiGPC using the well-know matrix inversion lemma \cite{bernstein2005matrix} (Woodbury identity) which provides a simple batched iterative rank-$B$ correction to the inverse of a given invertible matrix. 

\begin{table}[t]
    \centering
    \caption{{\bf Complexity Comparison of GP updates.}We observe a {\bf $\times$7.5} speedup in practice.}
    \begin{tabular}{cc}
    \hline
        \textbf{Classic GP update} & \textbf{Efficient GP update} \\\\[-1.5ex] 
        $\mathcal{O}(N_Q^3 + B N_Q^2)$ & $\mathcal{O}(B^3 + B N_Q^2 + B^2 N_Q)$ \\ \hline
    \end{tabular}
    \label{tab:cost}
\end{table}

In particular, for each labeled training mini-batch of size $B$, we replace the $B$ oldest samples in the buffer with the features computed using the current mini-batch. 
Let $K_{t-1}$ and $K_t$ be the covariance matrices at iteration $t-1$ and $t$ respectively. We now show how to compute $K_t$ by updating $K_{t-1}$ after having updated the memory bank with the new samples from the current mini-batch. Let, 
\begin{equation}
    K_{t}
    =
    \begin{bmatrix}k(h_o, h_o) & k(h_o, h_n)
    \\ k(h_o, h_n)^\top & k(h_n, h_n) \end{bmatrix} + \sigma^2 I = \begin{bmatrix} A& C\\ C^\top & D \end{bmatrix} \nonumber,
\end{equation}
where $h_o$ and $h_n$ denote the old samples that were kept in the buffer and the new samples added to the buffer. $I$ is the identity matrix. The inverse of $K_t$ is given by 
\begin{align}\centering
    K^{-1}_t &=  \begin{bmatrix} A& C\\ C^\top & D \end{bmatrix} ^{-1} 
    =\begin{bmatrix} K_{11}& K_{12}\\ K_{12}^\top & K_{22} \end{bmatrix} \nonumber \\
\text{where }    K_{22} &= (D -  C^\top A^{-1} C)^{-1} \nonumber \\
    K_{12} = - A^{-1} &C K_{22} \text{ and } K_{11} = A^{-1} +  A^{-1} C K_{22}C^\top A^{-1} \nonumber
\end{align}
Note that computing $K^{-1}_t$ only requires inverting the two matrices $A$ and $(D -  C^\top A^{-1} C)$. The latter is of size $B \times B$ while the former is still a relatively large matrix of size $(N_q - B) \times (N_q - B)$. However, $A$ does not depend on the newly added samples, yet it is not equal to $K^{-1}_{t-1}$. Therefore, $A^{-1}$ can be computed efficiently only requiring the inverse of a $B \times B$ matrix as follows:
\begin{equation}\label{eq:A_inv}
    A^{-1} = M_{11} - M_{12} M_{22}^{-1} M_{12}^\top; \quad K_{t-1}^{-1} =\begin{bmatrix} M_{11}& M_{12}\\ M_{12}^\top & M_{22} \end{bmatrix} \nonumber
\end{equation}
For simplicity, we assume that the new samples are located at the end of the buffer, however the derivation remains true for an arbitrarily ordered buffer up to a permutation matrix.

To summarize, using block matrix linear algebra, the cost of the inverting $K_t$ can be reduced to computing the inverse of a couple of $B \times B$ matrices which is in turn much more efficient when $B \ll N_q$ as is the case in our setting. The detailed derivation is provided in the supplementary material. For example, for a buffer size $N_Q \sim 16k$ and a batch size $B=8$, using our efficient update rule results in $\times 7.5$ speedup.

\noindent \textbf{Class-balanced SemiGPC.}
SemiGPC has the additional benefit of allowing us to explicitly address the class imbalance without altering the training scheme. We split $h_Q$ into $C$ class buffers and insert the new samples based on their labels, thus ensuring a balanced $h_Q$. We compare this approach to the classic class rebalancing in the supplementary material.

\section{Experimental Settings}

\subsection{Implementation details}\label{sec:usb}

We use the semi-supervised training recipe of USB~\cite{usb2022}\footnote{ \url{https://github.com/microsoft/semi-supervised-learning}}. It uses an ImageNet~\cite{deng2009imagenet} pre-trained ViT to initialize the student model. This training scheme allows for faster training time and better performance overall. We use a ViT Small/Tiny with a patch size of 2 and a resolution of 32 for CIFAR100/10 respectively. For our other experiments, we use a ViT Small with a patch size of 16 and a resolution of 224. All our experiments can be ran on a single V100 GPU. All our model are trained using AdamW~\cite{loshchilov2018adamw} for 200 epochs using a batch size of 8. The detailed set of hyper-parameters are provided in the supplementary material. For most of our experiments, we use SimMatch as our baseline. We include a comparison of SemiGPC across different algorithms in section~\ref{sec:algo}.
For all SemiGPC experiments, we use a buffer size $N_q=16300$. 
Following most works in the literature, we adopt the Top~1 Accuracy as our main evaluation metric and report the mean and standard deviation across 3 random seeds.

\subsection{Datasets}

\noindent \textbf{CIFAR10-LT, CIFAR100-LT.} We evaluate SemiGPC on imbalanced versions of CIFAR10 and CIFAR100. The class distribution of these datasets can be fully described using the imbalance ratio $\gamma$ and the number of samples in the majority class $N_1$. For each class $ 1 <i \leq K$ its number of samples $N_i$ is defined as 
\begin{equation}
    N_i = N_1 \gamma^{-\frac{i - 1}{K-1}}, \gamma = \nicefrac{N_1}{N_K}
\end{equation}
 where $\gamma$, $N_K$, and $K$ are the imbalance ratio, the cardinality of the minority class and the number of classes respectively. 

\noindent \textbf{FGVC Benchmarks.} We also evaluate SemiGPC on the fine-grained semi-supervised benchmarks introduces in~\cite{su2021realistic, su2021tax}. These challenging benchmarks contain naturally long-tailed distributions with highly similar class pair. Note that both works~\cite{su2021realistic, su2021tax} argue that Semi-supervised methods struggle on such benchmarks.
These datasets include a labeled set $L_{in}$ and two unlabeled $U_{in}$ and $U_{out}$ with seen and unseen classes.

\noindent \textbf{SemiAves.} This dataset~\cite{su2021realistic} is built using the Aves kingdom in iNaturalist 2018 dataset~\cite{inaturalist18}. $L_{in}$, $U_{in}$ and $U_{out}$ include 200/200/800 species and 5959/26640/122208 images respectively. The test set is balanced and contains 40 samples per class. Its reported imbalance ratio is $\gamma = 7.9$.

\noindent \textbf{SemiFungi.} This dataset is based on the CVPR 2018 FGVCx Fungi challenge dataset~\cite{semifungi}. $L_{in}$, $U_{in}$ and $U_{out}$ include 200/200/1194 species and 4141/13166/64871 images respectively.
The test set is balanced and contains 20 samples per class. Its reported imbalance ratio is $\gamma = 10.1$.

\noindent \textbf{Semi-iNat.} This dataset was introduced at CVPR 2021 FGVC8 workshop~\cite{su2021semiinat}. 
$L_{in}$, $U_{in}$ and $U_{out}$ include 810/810/1629 species and 13771/91336/221912 images respectively.
The test set is balanced and contains 100 samples per seen class. Its imbalance ratio is $\gamma = 8.5$.

\noindent \textbf{SemiCUB.} This dataset is based on the Caltech-UCSD Birds-200-2011 (CUB) dataset~\cite{cub200}. 
$L_{in}$, $U_{in}$ and $U_{out}$ include 100/100/100 species and 1000/3853/5903 images respectively.
Unlike the other three, only the unlabeled sets are imbalanced with $\gamma \sim 2$ for $U_{in}$.
The test set is balanced and contains 1000 samples.

We use $U = U_{in}$ as our unlabeled dataset. Results for $U = U_{in} \cup U_{out}$ are shown in the supplementary material.


\section{Experimental Results} 


In this section, we showcase SemiGPC's robustness under various degrees of class imbalance across different data regimes on CIFAR10-LT and CIFAR100-LT. We then report the performance on the more challenging long-tailed semi-supervised benchmarks SemiAves, SemiCUB, SemiFungi and Semi-iNat. Lastly, we benchmark SemiGPC on the classic balanced semi-supervised splits of CIFAR10 and CIFAR100. 
For all our imbalanced experiments, we forgo using techniques such as CReST as they don't necessarily improve performance when combined with the USB~\cite{usb2022} training recipe. These results can be found in the supplementary material.

\subsection{Imbalanced Semi-Supervised Learning}


In this section, we evaluate the robustness of SemiGPC under different degrees of class imbalance on CIFAR10-LT and CIFAR100-LT. More specifically, we explore two imbalanced settings based on whether one has access to a balanced labeled dataset or not:

\begin{itemize}
\item \textbf{Setting A ($\gamma_l = \gamma_u > 1$).} Both the labeled and unlabeled sets are imbalanced using  the same factor. Following prior works, we use $(N_1^l, N_1^u) = (150, 500)$ for the imbalanced version CIFAR100, \ie CIFAR100-LT, and $(N_1^l, N_1^u) = (1500, 5000)$ for the imbalanced version CIFAR10, \ie CIFAR10-LT. $N_1^l$ and $N_1^u$ are the number of samples for the majority class in the labeled and unlabeled datasets respectively.

\item \textbf{Setting B ($\gamma_l = 1; \gamma_u > 1$).} Only the unlabeled set is imbalanced. For the labeled setting, we use 4 samples per class resulting $\times100/\times10$ fewer labeled samples compared to A for CIFAR10-LT and CIFAR100-LT respectively. We argue that this setting is more challenging and better represents real-world scenarios. Indeed, realistically, a small set of balanced labeled samples can be curated while one cannot make any assumptions on the distribution of the unlabeled dataset based on its labeled counterpart.
\end{itemize}

\begin{table}[t]
\centering
\caption{Top1 Accuracy obtained on CIFAR100-LT for different values of  $\gamma_l = 1$ and $\gamma_u$. \dag: A class balanced buffer is used for SemiGPC. \textasteriskcentered: as reported by~\cite{fan2022CoSSL}. The difference to the baseline is highlighted in green/red.}
\label{tab:c100-lt}
\begin{tabular}{lcccl}
\hline
\textbf{\textbf{\textbf{Model}}}               & \textbf{\textbf{$\gamma_l$}} & \textbf{\textbf{\textbf{$\gamma_u$}}} & \textbf{$n_{lb}$} & \textbf{\textbf{Top1 Acc}}                                     \\ \hline
CoSSL~\cite{fan2022CoSSL}\textasteriskcentered & 20                           & 20                                    & 4741              & 55.80 {\tiny $\pm  0.62$}                                      \\
SimMatch                                       & 20                           & 20                                    & 4741              & 83.38 {\tiny $\pm  0.48$}                                      \\
w/ SemiGPC\dag                                 & 20                           & 20                                    & 4741              & \textbf{83.76 {\tiny $\pm  0.26$} {\bf \color{Green} (+0.37)}} \\ \hline
CoSSL~\cite{fan2022CoSSL}\textasteriskcentered & 50                           & 50                                    & 3751              & 48.90 {\tiny $\pm  0.61$}                                      \\
SimMatch                                       & 50                           & 50                                    & 3751              & 78.82 {\tiny $\pm  0.60$}                                      \\
w/ SemiGPC\dag                                 & 50                           & 50                                    & 3751              & \textbf{79.79 {\tiny $\pm  0.08$} {\color{Green} (+0.97)}}     \\ \hline
CoSSL~\cite{fan2022CoSSL}\textasteriskcentered & 100                          & 100                                   & 3218              & 44.10 {\tiny $\pm  0.59$}                                      \\
SimMatch                                       & 100                          & 100                                   & 3218              & 73.90 {\tiny $\pm  0.75$}                                      \\
w/ SemiGPC\dag                                 & 100                          & 100                                   & 3218              & \textbf{74.48 {\tiny $\pm  0.98$} {\color{Green} (+0.58)}}     \\ \hline \hline
SimMatch                                       & 1                            & 20                                    & 400               & 76.28 {\tiny $\pm  0.28$}                                      \\
w/ SemiGPC                                     & 1                            & 20                                    & 400               & \textbf{77.79 {\tiny $\pm  0.51$} {\color{Green} (+1.53)}}     \\ \hline
SimMatch                                       & 1                            & 50                                    & 400               & 72.78 {\tiny $\pm  0.29$}                                      \\
w/ SemiGPC                                     & 1                            & 50                                    & 400               & \textbf{75.21 {\tiny $\pm  0.53$} {\color{Green} (+2.43)}}     \\ \hline
SimMatch                                       & 1                            & 100                                   & 400               & 70.19 {\tiny $\pm  0.43$}                                      \\
w/ SemiGPC                                     & 1                            & 100                                   & 400               & \textbf{73.47 {\tiny $\pm  0.63$} {\color{Green} (+3.28)}}     \\ \hline
\end{tabular}
\end{table}

\noindent \textbf{CIFAR100-LT (Table \ref{tab:c100-lt}).}
For Setting A, we use the class-balanced version of SemiGPC. We also include the numbers reported by \cite{fan2022CoSSL} for CoSSL+ReMixMatch as they represent the current state of the art.
For setting A, 
we observe that SemiGPC outperforms the baseline across all values of $\gamma_u$.
This highlights SemiGPC's robustness with respect to class imbalance and its inherent ability to address it explicitly using a balanced buffer. 
The results obtained for setting B further support the robustness of SemiGPC with respect to class imbalance. Indeed, when provided with balanced samples that are $10\times$ fewer than setting A, SemiGPC is able to outperform our baseline across all values of $\gamma_u$ by a margin greater than $+1.5\%$.
Additionally, for each model and value $\gamma_u$ we measure the gap $\Delta(\gamma_u) = \text{Acc}(A) - \text{Acc}(B)$ between the accuracies $\text{Acc}(A)$ and $\text{Acc}(b)$ in setting A and B respectively.
When comparing $\Delta$ averaged over all $\gamma_u$ values, we observe a gap of $5.62\%$ and $3.85\%$ for SimMatch and SemiGPC respectively. In addition to improving performance across both settings, SemiGPC is better at bridging the gap between the two data regimes by about $32\%$.

\noindent \textbf{CIFAR10-LT (Table~\ref{tab:c10-lt}).}
For setting A, we observe the SemiGPC outperforms the baseline across different values of $\gamma_u$ especially for the more challenging setting $\gamma_u=150$ where we observe a gap of $+1.34\%$. This highlights the robustness of SemiGPC to class imbalance.
SemiGPC also largely outperforms our baseline in the setting B. We observe an accuracy increase of at least $7.63\%$ across all values of $\gamma_u$ with the gap growing bigger for higher values of $\gamma_u$ up to $+10.17\%$.
Additionally, we report an average gap across settings of $17.39\%$ and $9.54\%$ for SimMatch and SemiGPC respectively, \ie, a relative improvement of $45\%$. Thanks to its normalization scheme, SemiGPC reduces the risk of confirmation bias which is more prominent when the labeled data is scarce.
\begin{table}[t!]
\centering
\caption{Top1 Accuracy obtained on CIFAR10-LT for different values of  $\gamma_l = 1$ and $\gamma_u$. \dag: A class balanced buffer is used for SemiGPC. \textasteriskcentered: as reported by~\cite{fan2022CoSSL}.  The difference to the baseline is highlighted in green.}
\label{tab:c10-lt}
\begin{tabular}{lcccl}
\hline
\textbf{\textbf{\textbf{Model}}}               & \textbf{\textbf{$\gamma_l$}} & \textbf{\textbf{$\gamma_u$}} & \textbf{\textbf{$n_{lb}$}} & \textbf{\textbf{Top1 Acc}}                                      \\ \hline
CoSSL~\cite{fan2022CoSSL}\textasteriskcentered & 50                           & 50                           & 4196                       & 87.70 {\tiny $\pm 0.21$}                                        \\
SimMatch                                       & 50                           & 50                           & 4196                       & 96.48 {\tiny $\pm 0.26$}                                        \\
w/ SemiGPC\dag                                 & 50                           & 50                           & 4196                       & \textbf{96.80 {\tiny $\pm 0.12$}   {\bf \color{Green} (+0.32)}} \\ \hline
CoSSL~\cite{fan2022CoSSL}\textasteriskcentered & 100                          & 100                          & 3720                       & 84.10 {\tiny $\pm 0.56$}                                        \\
SimMatch                                       & 100                          & 100                          & 3720                       & 94.59 {\tiny $\pm 0.59$}                                        \\
w/ SemiGPC\dag                                 & 100                          & 100                          & 3720                       & \textbf{95.74 {\tiny $\pm 0.37$}  {\color{Green} (+1.15)}}      \\ \hline
CoSSL~\cite{fan2022CoSSL}\textasteriskcentered & 150                          & 150                          & 3496                       & 81.30 {\tiny $\pm 0.83$}                                        \\
SimMatch                                       & 150                          & 150                          & 3496                       & 94.07 {\tiny $\pm 1.46$}                                        \\
w/ SemiGPC\dag                                 & 150                          & 150                          & 3496                       & \textbf {95.41 {\tiny $\pm 0.56$} {\color{Green} (+1.34)}}      \\ \hline \hline
SimMatch                                       & 1                            & 50                           & 40                         & 80.59 {\tiny $\pm 2.24$}                                        \\
w/ SemiGPC                                     & 1                            & 50                           & 40                         & \textbf{88.22 {\tiny $\pm 2.38$} {\color{Green} (+7.63)}}       \\ \hline
SimMatch                                       & 1                            & 100                          & 40                         & 76.69 {\tiny $\pm 2.13$}                                        \\
w/ SemiGPC                                     & 1                            & 100                          & 40                         & \textbf{86.86 {\tiny $\pm 4.48$} {\color{Green} (+10.17)}}      \\ \hline
SimMatch                                       & 1                            & 150                          & 40                         & 75.68 {\tiny $\pm 4.31$}                                        \\
w/ SemiGPC                                     & 1                            & 150                          & 40                         & \textbf{84.25 {\tiny $\pm 8.92$} {\color{Green} (+8.57)}}       \\ \hline
\end{tabular}
\end{table}

\subsection{Semi-Supervised FGVC Benchmarks}\label{sec:semix}

In the section, we evaluate the performance of SemiGPC on the naturally long-tailed semi-supervised benchmarks such as SemiAves, SemiFungi and SemiCUB and for Semi-iNat. In addition to the class imbalance, these datasets includes highly similar classes, \cf supplementary material.

\begin{table}
\centering
\caption{Top1 Accuracy obtained on the considered fine-grained semi-supervised benchmarks.\textasteriskcentered: as reported by ~\cite{su2021realistic, su2021tax}. The difference to the baseline is highlighted in green.}
\label{tab:semiX}
\begin{tabular}{lll}
\hline
\multicolumn{1}{c}{\textbf{Dataset}} & \multicolumn{1}{c}{\textbf{Model}}         & \textbf{\textbf{Top1 Acc}}                                \\ \hline
\multirow{3}{*}{SemiCUB}             & FixMatch~\cite{su2021realistic}\textasteriskcentered & 53.20                                                     \\
                                     & SimMatch                                   & 84.53 {\tiny $\pm 0.45$}                                  \\
                                     & w/ SemiGPC                        & \textbf{85.43 {\tiny $\pm 0.67$} {\color{Green} (+0.90)}} \\ \hline
\multirow{3}{*}{SemiAves}            & FixMatch~\cite{su2021realistic}\textasteriskcentered & 57.40 {\tiny $\pm 0.80$}                                  \\
                                     & SimMatch                                   & 68.47 {\tiny $\pm 0.43$}                                  \\
                                     & w/ SemiGPC                          & \textbf{69.59 {\tiny $\pm 0.09$} {\color{Green} (+1.12)}} \\ \hline
\multirow{3}{*}{SemiFungi}           & FixMatch~\cite{su2021realistic}\textasteriskcentered & 56.30 {\tiny $\pm 0.50$}                                  \\
                                     & SimMatch                                   & 68.01{\tiny $\pm 0.19$}                                   \\
                                     & w/ SemiGPC                          & \textbf{71.50 {\tiny $\pm 0.49$} {\color{Green} (+3.49)}} \\ \hline
\multirow{3}{*}{Semi-iNat}           & FixMatch~\cite{su2021tax}\textasteriskcentered                  & 44.10                                                     \\
                                     & SimMatch                                   & 64.95 {\tiny $\pm 0.11$}                                  \\
                                     & w/ SemiGPC                          & \textbf{66.54 {\tiny $\pm 0.85$} {\color{Green} (+1.59)}} \\ \hline 
\end{tabular}
\end{table}

We report the obtained results on Table~\ref{tab:semiX}. For reference, we report the numbers obtained by~\cite{su2021realistic} for SemiAves, SemiFungi and SemiCUB and by~\cite{su2021tax} for Semi-iNat. We show that using the USB~\cite{usb2022} training recipe produces a strong semi-supervised baseline as opposed to the numbers reported by~\cite{su2021realistic, su2021tax}. Furthermore, SemiGPC outperforms the baseline on all fine-grained benchmarks. This is especially true for the most imbalanced dataset SemiFungi ($\gamma=10.1$) where SemiGPC improves upon the baseline accuracy by $+3.49\%$. This shows that SemiGPC is not only more robust with respect to class imbalance on artificially skewed benchmarks such CIFAR10/100-LT but is  also better suited for naturally imbalanced datasets containing fine-grained classes where it  establishes a new state of the art.

\subsection{Standard CIFAR10/CIFAR100}

Lastly, we evaluate our SemiGPC method on different split of CIFAR100 and CIFAR10. We report the obtained performance on Table~\ref{tab:c100} when using 200/400 and 40/250 labeled samples for CIFAR100 and CIFAR10 respectively. For reference we include the numbers reported by USB~\cite{usb2022} and FreeMatch~\cite{wang2023freematch} as the current state of the art. We observe that SemiGPC improves performance across different amount of available labeled samples on CIFAR100, with the biggest improvement $+0.83\%$ in the low data regime. However, SemiGPC is simply on par with the baseline on CIFAR10. We argue that the semi-supervised performance is already saturated on this benchmark when using the USB training recipe.


\begin{table}[t]
\centering
\caption{Top1 Accuracy obtained on CIFAR10 and CIFAR100 for different numbers of labeled samples. \textasteriskcentered: reported by~\cite{usb2022,wang2023freematch}.}
\label{tab:c100}
\label{tab:my-table}
\begin{tabular}{llcl}
\hline
Dataset  & \textbf{Model} & \textbf{$n_{lb}$} & \textbf{Top1 Acc}                       \\ \hline
CIFAR100 & USB~\cite{usb2022}\textasteriskcentered     & 200               & 79.15                                   \\
CIFAR100 & SimMatch       & 200               & 79.18                                   \\
CIFAR100 & w/ SemiGPC     & 200               & \textbf{80.01 {\color{Green} (+0.83)} } \\ \hline
CIFAR100 & USB~\cite{usb2022}\textasteriskcentered     & 400               & 83.20                                   \\
CIFAR100 & SimMatch       & 400               & 83.25                                   \\
CIFAR100 & w/ SemiGPC     & 400               & \textbf{83.87 {\color{Green} (+0.62)}}  \\ \hline
CIFAR10  & FreeMatch~\cite{wang2023freematch}\textasteriskcentered   & 40            & 95.10                         \\
CIFAR10  & SimMatch       & 40                & \textbf{97.32}                          \\
CIFAR10  & w/ SemiGPC     & 40                & 97.14 {\bf \color{Red} (-0.18)}         \\ \hline
CIFAR10  & FreeMatch~\cite{wang2023freematch}\textasteriskcentered       & 250       & 95.12                                   \\
CIFAR10  & SimMatch       & 250               & 97.21                                   \\
CIFAR10  & w/ SemiGPC     & 250               & \textbf{97.39 {\color{Green} (-0.18)}}  \\ \hline
\end{tabular}
\end{table}

\section{Ablations}

In order to establish the general purpose nature of SemiGPC, throughout this section we highlight the impact of SemiGPC on top of different underlying algorithms and/or pre-training strategies.

\subsection{Semi-Supervised Learning Algorithms}\label{sec:algo}

The design of SemiGPC is agnostic to the underlying choice of the semi-supervised algorithms.
In this section, we evaluate the impact of our proposed GP-based classifier on different Semi-Supervised methods including FixMatch~\cite{sohn2020fixmatch}, ReMixMatch~\cite{berthelot2020remixmatch}, SimMatch~\cite{zheng2022simmatch} and FreeMatch~\cite{wang2023freematch} on the SemiAves benchmark. The obtained results are reported in Table~\ref{tab:algo}. 
Despite FreeMatch~\cite{wang2023freematch} being designed to better tackle class imbalance, we observe that SimMatch~\cite{zheng2022simmatch} outperforms it when using the USB~\cite{usb2022} training recipe. This justifies why we use SimMatch as our baseline throughout this work.
\begin{table}[t]
\centering
\caption{Comparison of the Top1 Accuracy on SemiAves when using different Semi-supervised algorithms.}
\label{tab:algo}
\begin{tabular}{lcl}
\hline
\textbf{Model}   & \textbf{Dataset} & \textbf{Top1 Acc} \\ \hline
FreeMatch        & SemiAves         & 66.97             \\
w/ SemiGPC    & SemiAves         & \textbf{67.93 {\color{Green} (+0.96)}}    \\ \hline
FixMatch         & SemiAves         & 67.36             \\
w/ SemiGPC     & SemiAves         & \textbf{68.31 {\color{Green} (+0.95)}}    \\ \hline
ReMixMatch       & SemiAves         & 67.9              \\
w/ SemiGPC & SemiAves         & \textbf{68.31 {\color{Green} (+0.41)}}    \\ \hline
SimMatch         & SemiAves         & 68.45             \\
w/ SemiGPC     & SemiAves         & \textbf{69.30 {\color{Green} (+0.85)}}    \\ \hline
\end{tabular}
\end{table}
Not only does SemiGPC improve performance across all considered methods, but its performance also improves monotonically with respect to the performance of the base method. This allows SemiGPC to remain relevant to future better semi-supervised algorithms.

\subsection{Pre-training Strategy}\label{sec:pretrain}

As stated in section~\ref{sec:usb}, we used a pre-trained ViT~\cite{dosovitskiy2020vit} to initialize our semi-supervised models. We evaluate the impact of SemiGPC across different pre-training strategies by training SimMatch on the SemiAves benchmark using supervised pre-training, Dino~\cite{caron2021dino} and MSN~\cite{assran2022msn} pre-training on ImageNet~\cite{deng2009imagenet}. Both Dino~\cite{caron2021dino} and MSN~\cite{assran2022msn} are self-supervised methods that produce competitive performance on ImageNet with MSN being the top performer out of the two. We report the obtained results in Table~\ref{tab:pretrain}. Not only does the SemiGPC performance scale based on the performance of the pre-training methods, it also improves performance across all considered pre-training strategies. 

\begin{table}[t]
\centering
\caption{Comparison of the Top1 Accuracy on SemiAves when using different pre-training strategies.}
\label{tab:pretrain}
\begin{tabular}{lcl}
\hline
\textbf{Model} & \textbf{Pretraining} & \textbf{Top1 Acc} \\ \hline
SimMatch       & DINO                 & 64                \\
w/ SemiGPC   & DINO                 & \textbf{65.32 {\color{Green} (+1.32)}}    \\ \hline
SimMatch       & MSN                  & 64.7              \\
w/ SemiGPC   & MSN                  & \textbf{67.73 {\color{Green} (+3.03)}}    \\ \hline
SimMatch       & Supervised                  & 68.45             \\
w/ SemiGPC   & Supervised                  & \textbf{69.30 {\color{Green} (+0.85)}}    \\ \hline
\end{tabular}
\end{table}
\section{Conclusion and Future Work}

Our method SemiGPC is able to achieve state of the art results across different benchmarks and settings thanks to its ability to counteract imbalances in the data distribution. However, SemiGPC still has a few limitations.
We observe in Tables~\ref{tab:c10-lt} and \ref{tab:c100} that SemiGPC shows mixed results when used on top of an already strong baseline ($>94\%$ accuracy) such as on CIFAR10. 
Also, although our update rule greatly speeds up the matrix inversion, it does not fully eliminate the additional computational overhead.  Furthermore, the quadratic scaling of the memory cost of this matrix limits the maximum size of the buffer in SemiGPC to around $N_Q = 16K$. However, this limitation can be addressed using an ensemble of GPs each using a separate buffer. This would allow us to scale SemiGPC to $N_Q \sim 80K$. 
Furthermore, our update rule is not compatible with using trainable kernel hyper-parameters since it relies on reusing previous values of the kernel matrix. Leveraging matrix-vector-matrix solvers~\cite{wang2019exact} fixes both these limitations. Indeed, by enabling efficient GP inference with trainable hyper-parameters, SemiGPC would forgo sharing the kernel hyper-parameters across classes and adapt the geometry induced by the kernel function on a per-class basis.
We leave deriving an online update rule using matrix-vector-matrix solvers to future work. 
Lastly, combining SemiGPC with alternative definitions of confidence to eq.~\eqref{eq:conf} by either using the sample-wise posterior covariance provided by GP or by leveraging recent advances in efficient Neural Tangent Kernel (NTK) computation~\cite{novak2022fast, zhou2021meta} remains an open area of research.
{\small
\bibliographystyle{ieee_fullname}
\bibliography{egbib}

\begin{thebibliography}{10}\itemsep=-1pt

\bibitem{semifungi}
2018 fgvcx fungi classification challenge.
\newblock \url{https://github.com/visipedia/fgvcx_fungi_comp}.

\bibitem{assran2022msn}
Mahmoud Assran, Mathilde Caron, Ishan Misra, Piotr Bojanowski, Florian Bordes, Pascal Vincent, Armand Joulin, Mike Rabbat, and Nicolas Ballas.
\newblock Masked siamese networks for label-efficient learning.
\newblock In {\em Computer Vision--ECCV 2022: 17th European Conference, Tel Aviv, Israel, October 23--27, 2022, Proceedings, Part XXXI}, pages 456--473. Springer, 2022.

\bibitem{bernstein2005matrix}
D Bernstein.
\newblock Matrix mathematics (princeton university press.
\newblock page~45, 2005.

\bibitem{berthelot2020remixmatch}
David Berthelot, Nicholas Carlini, Ekin~Dogus Cubuk, Alexey Kurakin, Kihyuk Sohn, Han Zhang, and Colin Raffel.
\newblock Remixmatch: Semi-supervised learning with distribution matching and augmentation anchoring.
\newblock In {\em International Conference on Learning Representations}, 2020.

\bibitem{caron2021dino}
Mathilde Caron, Hugo Touvron, Ishan Misra, Herv{\'e} J{\'e}gou, Julien Mairal, Piotr Bojanowski, and Armand Joulin.
\newblock Emerging properties in self-supervised vision transformers.
\newblock In {\em Proceedings of the IEEE/CVF international conference on computer vision}, pages 9650--9660, 2021.

\bibitem{cubuk2019autoaugment}
Ekin~D Cubuk, Barret Zoph, Dandelion Mane, Vijay Vasudevan, and Quoc~V Le.
\newblock Autoaugment: Learning augmentation strategies from data.
\newblock In {\em Proceedings of the IEEE/CVF conference on computer vision and pattern recognition}, pages 113--123, 2019.

\bibitem{deng2009imagenet}
Jia Deng, Wei Dong, Richard Socher, Li-Jia Li, Kai Li, and Li Fei-Fei.
\newblock Imagenet: A large-scale hierarchical image database.
\newblock In {\em 2009 IEEE conference on computer vision and pattern recognition}, pages 248--255. Ieee, 2009.

\bibitem{devries2017cutout}
Terrance DeVries and Graham~W Taylor.
\newblock Improved regularization of convolutional neural networks with cutout.
\newblock {\em arXiv preprint arXiv:1708.04552}, 2017.

\bibitem{dosovitskiy2020vit}
Alexey Dosovitskiy, Lucas Beyer, Alexander Kolesnikov, Dirk Weissenborn, Xiaohua Zhai, Thomas Unterthiner, Mostafa Dehghani, Matthias Minderer, Georg Heigold, Sylvain Gelly, Jakob Uszkoreit, and Neil Houlsby.
\newblock An image is worth 16x16 words: Transformers for image recognition at scale.
\newblock {\em ICLR}, 2021.

\bibitem{dutordoir2021deep}
Vincent Dutordoir, James Hensman, Mark van~der Wilk, Carl~Henrik Ek, Zoubin Ghahramani, and Nicolas Durrande.
\newblock Deep neural networks as point estimates for deep gaussian processes.
\newblock In {\em NeurIPS}, 2021.

\bibitem{fan2022CoSSL}
Yue Fan, Dengxin Dai, Anna Kukleva, and Bernt Schiele.
\newblock Cossl: Co-learning of representation and classifier for imbalanced semi-supervised learning.
\newblock In {\em Proceedings of the IEEE/CVF conference on computer vision and pattern recognition}, pages 14574--14584, 2022.

\bibitem{gal2016dropout}
Yarin Gal and Zoubin Ghahramani.
\newblock Dropout as a bayesian approximation: Representing model uncertainty in deep learning.
\newblock In {\em international conference on machine learning}, pages 1050--1059. PMLR, 2016.

\bibitem{inaturalist18}
{iNaturalist} 2018 competition dataset.
\newblock ~\url{https://github.com/visipedia/inat_comp/tree/master/2018}, 2018.

\bibitem{johnander2022dense}
Joakim Johnander, Johan Edstedt, Michael Felsberg, Fahad~Shahbaz Khan, and Martin Danelljan.
\newblock Dense gaussian processes for few-shot segmentation.
\newblock In {\em ECCV}, 2022.

\bibitem{lawrence2004semi}
Neil Lawrence and Michael Jordan.
\newblock Semi-supervised learning via gaussian processes.
\newblock {\em Advances in neural information processing systems}, 17, 2004.

\bibitem{pmlr-v164-lee22c}
Jongseok Lee, Jianxiang Feng, Matthias Humt, Marcus~Gerhard M\"uller, and Rudolph Triebel.
\newblock Trust your robots! predictive uncertainty estimation of neural networks with sparse gaussian processes.
\newblock In {\em CoRL}, 2022.

\bibitem{li2021comatch}
Junnan Li, Caiming Xiong, and Steven~CH Hoi.
\newblock Comatch: Semi-supervised learning with contrastive graph regularization.
\newblock In {\em Proceedings of the IEEE/CVF international conference on computer vision}, pages 9475--9484, 2021.

\bibitem{liu2020uncertainty}
Zhao-Yang Liu, Shao-Yuan Li, Songcan Chen, Yao Hu, and Sheng-Jun Huang.
\newblock Uncertainty aware graph gaussian process for semi-supervised learning.
\newblock In {\em Proceedings of the AAAI Conference on Artificial Intelligence}, volume~34, pages 4957--4964, 2020.

\bibitem{loshchilov2018adamw}
Ilya Loshchilov and Frank Hutter.
\newblock Decoupled weight decay regularization.
\newblock In {\em International Conference on Learning Representations}, 2019.

\bibitem{novak2022fast}
Roman Novak, Jascha Sohl-Dickstein, and Samuel~S Schoenholz.
\newblock Fast finite width neural tangent kernel.
\newblock In {\em International Conference on Machine Learning}, pages 17018--17044. PMLR, 2022.

\bibitem{rasmussen2006gaussian}
Carl~Edward Rasmussen, Christopher~KI Williams, et~al.
\newblock {\em Gaussian processes for machine learning}, volume~1.
\newblock Springer, 2006.

\bibitem{sindhwani2007semi}
Vikas Sindhwani, Wei Chu, and S~Sathiya Keerthi.
\newblock Semi-supervised gaussian process classifiers.
\newblock In {\em IJCAI}, pages 1059--1064, 2007.

\bibitem{sohn2020fixmatch}
Kihyuk Sohn, David Berthelot, Nicholas Carlini, Zizhao Zhang, Han Zhang, Colin~A Raffel, Ekin~Dogus Cubuk, Alexey Kurakin, and Chun-Liang Li.
\newblock Fixmatch: Simplifying semi-supervised learning with consistency and confidence.
\newblock {\em Advances in neural information processing systems}, 33:596--608, 2020.

\bibitem{su2021realistic}
Jong-Chyi Su, Zezhou Cheng, and Subhransu Maji.
\newblock A realistic evaluation of semi-supervised learning for fine-grained classification.
\newblock In {\em Proceedings of the IEEE/CVF Conference on Computer Vision and Pattern Recognition}, pages 12966--12975, 2021.

\bibitem{su2021semiinat}
Jong-Chyi Su and Subhransu Maji.
\newblock The semi-supervised inaturalist challenge at the fgvc8 workshop, 2021.

\bibitem{su2021tax}
Jong-Chyi Su and Subhransu Maji.
\newblock Semi-supervised learning with taxonomic labels.
\newblock In {\em British Machine Vision Conference (BMVC), 2021}, 2021.

\bibitem{cub200}
C. Wah, S. Branson, P. Welinder, P. Perona, and S. Belongie.
\newblock The caltech-ucsd birds-200-2011 dataset.
\newblock Technical Report CNS-TR-2011-001, California Institute of Technology, 2011.

\bibitem{wang2022np}
Jianfeng Wang, Thomas Lukasiewicz, Daniela Massiceti, Xiaolin Hu, Vladimir Pavlovic, and Alexandros Neophytou.
\newblock Np-match: When neural processes meet semi-supervised learning.
\newblock In {\em International Conference on Machine Learning}, pages 22919--22934. PMLR, 2022.

\bibitem{wang2019exact}
Ke Wang, Geoff Pleiss, Jacob Gardner, Stephen Tyree, Kilian~Q Weinberger, and Andrew~Gordon Wilson.
\newblock Exact gaussian processes on a million data points.
\newblock {\em Advances in neural information processing systems}, 32, 2019.

\bibitem{usb2022}
Yidong Wang, Hao Chen, Yue Fan, Wang Sun, Ran Tao, Wenxin Hou, Renjie Wang, Linyi Yang, Zhi Zhou, Lan-Zhe Guo, Heli Qi, Zhen Wu, Yu-Feng Li, Satoshi Nakamura, Wei Ye, Marios Savvides, Bhiksha Raj, Takahiro Shinozaki, Bernt Schiele, Jindong Wang, Xing Xie, and Yue Zhang.
\newblock Usb: A unified semi-supervised learning benchmark for classification.
\newblock In {\em Thirty-sixth Conference on Neural Information Processing Systems Datasets and Benchmarks Track}, 2022.

\bibitem{wang2023freematch}
Yidong Wang, Hao Chen, Qiang Heng, Wenxin Hou, Yue Fan, , Zhen Wu, Jindong Wang, Marios Savvides, Takahiro Shinozaki, Bhiksha Raj, Bernt Schiele, and Xing Xie.
\newblock Freematch: Self-adaptive thresholding for semi-supervised learning.
\newblock 2023.

\bibitem{yang2019wide}
Greg Yang.
\newblock Wide feedforward or recurrent neural networks of any architecture are gaussian processes.
\newblock In {\em NeurIPS}, 2019.

\bibitem{zheng2022simmatch}
Mingkai Zheng, Shan You, Lang Huang, Fei Wang, Chen Qian, and Chang Xu.
\newblock Simmatch: Semi-supervised learning with similarity matching.
\newblock In {\em Proceedings of the IEEE/CVF Conference on Computer Vision and Pattern Recognition}, pages 14471--14481, 2022.

\bibitem{zhou2021meta}
Yufan Zhou and Zhenyi Wang.
\newblock Meta-learning with neural tangent kernels.
\newblock In {\em The International Conference on Learning Representations (ICLR)}, 2021.

\end{thebibliography}
}

\end{document}